\documentclass[10pt,twocolumn,letterpaper]{article}

\usepackage[pagenumbers]{cvpr} 

\usepackage{graphicx}
\usepackage{blindtext}
\usepackage{amsmath}
\usepackage{amssymb}
\usepackage{booktabs}
\usepackage{multirow}
\usepackage[pagebackref,breaklinks,colorlinks]{hyperref}

\usepackage[capitalize]{cleveref}
\crefname{section}{Sec.}{Secs.}
\Crefname{section}{Section}{Sections}
\Crefname{table}{Table}{Tables}
\crefname{table}{Tab.}{Tabs.}
\UseRawInputEncoding

\def\cvprPaperID{*****} 
\def\confName{CVPR}
\def\confYear{2022}

\begin{document}

\title{{\fontsize{13.1pt}{0}\selectfont Industrial Style Transfer with Large-scale Geometric Warping and Content Preservation}}
\author{Jinchao Yang$^{1\ast}$,\ \ Fei Guo$^{1\ast}$,\ \ Shuo Chen$^2$,\ \ Jun Li$^{1\dag}$,\ \ Jian Yang$^1$ \\
$^1$PCA Lab, Nanjing University of Science and Technology \ \ \ \ \ \ $^2$RIKEN \\
{\tt \small \{yangjinchao,feiguo,junli,csjyang\}@njust.edu.cn \ \ \ \   shuo.chen.ya@riken.jp } \\
{\tt \small $^\ast$Contributes equally \ \ \ \ $^\dag$Corresponding author\&project lead}
}

\twocolumn[{%
\renewcommand\twocolumn[1][]{#1}%
\maketitle
\begin{center}
\vskip -0.3in
    \centering
    \captionsetup{type=figure}
    \includegraphics[width=0.97\textwidth,height=4.7cm]{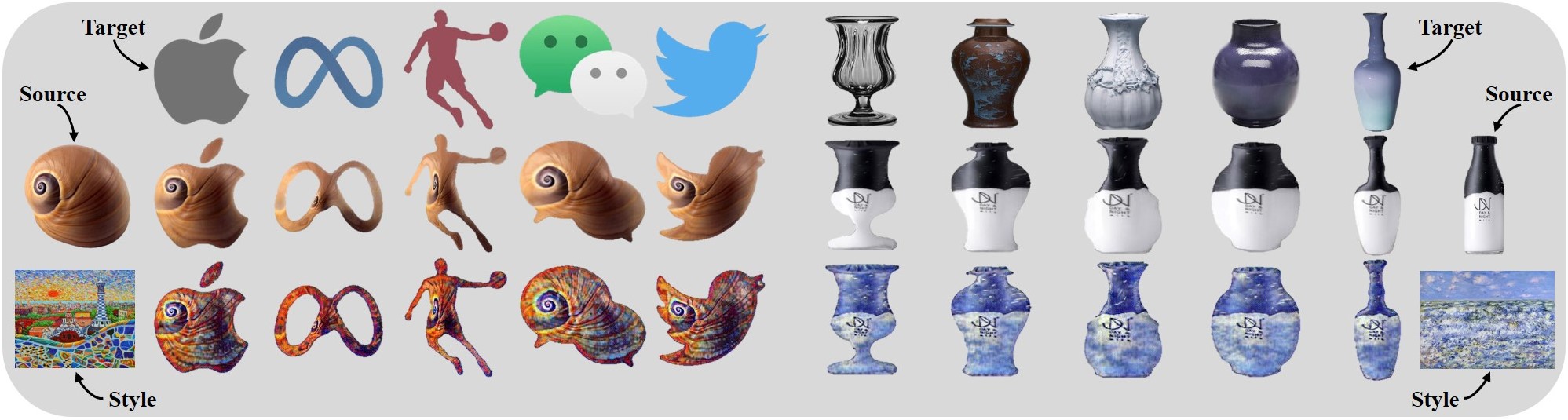}
    \vskip -0.1in
    \captionof{figure}{We propose an Industrial Style Transfer method for visual product design. Our method creates new product appearances (\eg, logos and Day$\&$Night bottles) by transferring both the shape of one product (target) and art style reference to another (source).}
    \label{fig:first}
\end{center}%
}]


\begin{abstract}
We propose a novel style transfer method to quickly create a new visual product with a nice appearance for industrial designersЎЇ reference. Given a source product, a target product, and an art style image, our method produces a neural warping field that warps the source shape to imitate the geometric style of the target and a neural texture transformation network that transfers the artistic style to the warped source product. Our model, Industrial Style Transfer (InST), consists of large-scale geometric warping (LGW) and interest-consistency texture transfer (ICTT). LGW aims to explore an unsupervised transformation between the shape masks of the source and target products for fitting large-scale shape warping. Furthermore, we introduce a mask smoothness regularization term to prevent the abrupt changes of the details of the source product. ICTT introduces an interest regularization term to maintain important contents of the warped product when it is stylized by using the art style image. Extensive experimental results demonstrate that InST achieves state-of-the-art performance on multiple visual product design tasks, \eg, companies' snail logos and classical bottles (please see Fig. \ref{fig:first}). To the best of our knowledge, we are the first to extend the neural style transfer method to create industrial product appearances. Code is available at \url{https://jcyang98.github.io/InST/home.html}
\end{abstract}

\section{Introduction}
\label{sec:intro}
Visual Product Design (VPD) has been recognized as a central role in the industrial product design field as consumers' choices heavily depend on the visual appearance of a new product in the marketplace \cite{Crilly2004pd}.
VPD often designs a novel product \cite{Creusen2005vpa} by following different appearance roles (\eg, aesthetic, functional, and symbolic). For example, designers usually produce the beautiful appearance of flying cars by referring to airplanes and cars to fuse their flying and driving functions and attractive aesthetic. However, it is difficult to quickly create high-quality product appearances due to the human intelligence in the VPD process, which is heavily dependent on designers' creative ability. Fortunately, neural style transfer (NST) \cite{Gatys2016imagestyletransfer,Huang2017adain,Kim2020DeformableST,Liu2021warp}, aiming at transferring artistic and geometric styles of one or two reference images to a content image, has a strong opportunity to assist the designers because the art style transformation is suitable for the aesthetic value, and some geometric shape transformations can gain the functional and symbolic values, such as Beijing National Stadium (bird's nest and building). Therefore, we seek for a style transfer formulation to automatically generate many visual appearance candidates of new products for industrial designers' reference.

However, most modern NST methods \cite{Li2019lineartransfer,Yang2019textimage_st,Deng2021video,Cheng2021normloss,Jiang2021fashion}, including geometric NST \cite{Kim2020DeformableST,Liu2021warp}, are difficultly or impossibly extended into directly design visual product appearances due to the following two challenges. \emph{One is the large-scale geometric shape between diverse objects (or products)} since designing new products is often to fuse two objects with very different geometries, such as flying cars (airplanes and cars) and butterfly doors (butterfly's wings and car doors). \emph{Another is that NST usually makes the content worse in the stylization process}, \eg, AdaIN \cite{Huang2017adain}, and WCT \cite{Li2017wct}, resulting in that product designers cannot refer to both rich content and novel geometry for generating creative inspirations.

\begin{figure}[t]
    \centering
    \includegraphics[width=1\linewidth]{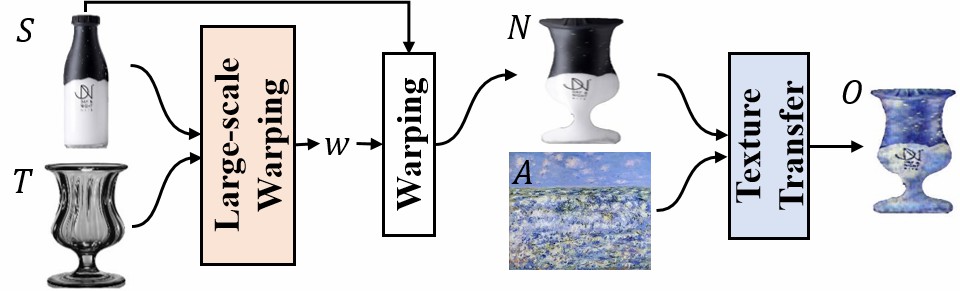}
    \vskip -0.1in
    \caption{The pipeline of our industrial style transfer. Our method creates a new product $N$ by warping source $S$ to target $T$, and generates a final product appearance $O$ by transferring the artistic style of reference image $A$ to the new product $N$. }
    \label{fig:overall}
    \vskip -0.15in
\end{figure}

To address these challenges, we develop an industrial style transfer (InST) method to create new product appearances, shown in Fig.~\ref{fig:overall}. Given a source product (or object), a target product, and an art reference image, InST aims to transfer both the industrially geometric shape of the target product and the art style of the reference image to the source product. In contrast to the existing NST methods, InST consists of large-scale geometric warping (LGW) and interest-consistency texture transfer (ICTT). Unlike small-scale geometric NST \cite{Kim2020DeformableST,Liu2021warp}, LGW designs a neural warping field between the shape masks of the source and target products using a shape-consistency loss. This is different from a warping field between their textural pixels as it results in a worse optimization, that is, a failure deformation. In addition, we explore a mask smoothness regularization term to prevent the abrupt changes of the details of the source product. With the help of the masks, LGW works well in large-scale warping between two products even though they are semantically irrelevant.

ICTT aims to keep interesting contents of the new product when it is stylized by using the art reference image. Inspired by the SuperPoint network \cite{Detone2018superpoint}, we present an interest regularization (IR) term based on both interest points and descriptors to constrain the art stylization to minimize perceptual differences between the new product and its stylized product. Unlike the most relevant work, ArtFlow \cite{An2021ArtFlow}, we design the interesting perceptual constraint to prevent worse contents, and our IR can further improve the performance of ArtFlow. Overall, the contributions of this work are summarized as follows:
\begin{itemize}
\item For large-scale geometric differences in the visual product design process, we explore a large-scale geometric warping module based on mask to transfer geometric shape style from one object product to another, even though irrelevant semantics.
\item For product content maintenance in the stylization process, we introduce an interest-consistency texture transfer with interesting regularization using interest point and descriptor extracted by the SuperPoint network to preserve content details.
\item Combining LGW with ICTT, we propose an industrial style transfer framework to fast generate visual appearances of new products, \eg, companies' logos, flying cars, and porcelain fashions. To the best of our knowledge, this work could open up a new field of style transfer, designing industrial product appearances.
\end{itemize}

\section{Related Work}
\label{sec:formatting}
In this section, we mainly review visual product design, texture style transfer, and geometric style transfer since we extend the style transfer technology to a new application, product appearance design tasks.
\subsection{Visual Product Design}
Due to the important influence of
product appearance on consumersЎЇ perceptions \cite{Bloch1995,Bloch2003}, visual product design (VPD) can be considered as a communication process between designers (companies) and consumers \cite{Crilly2004pd}. In this process, the designer aims to communicate a specific message through the product appearance by
changing the geometry, art style \etc, and the consumers provide their responses to the designers for the product improvements when seeing the product appearance \cite{Mugge2018vpd}. Usually, there are four popular types of product appearance for consumers: aesthetic impression, functional effect, symbolic association and ergonomic information \cite{Crilly2004pd,Creusen2005vpa,Radford2011vpd}.

However, this is a manual labour procedure with an expensive cost of communication and product design because it needs many feedback loops and designers take a lot of time to improve the product design in each loop \cite{Crilly2004pd}. This expensive cost drives us to explore a fast design method. More importantly, since high-quality product appearance is dependent on the designerЎЇs creative ability, it encourages us to generate many product innovations to inspire the designer. Therefore, we develop a novel style transfer method to create many visual product appearance candidates to assist or inspire designers.

\subsection{Texture Style Transfer.} As a hot topic, texture style transfer has developed for a long time. The initial works \cite{Gatys2016imagestyletransfer,Gatys2016preserving,Li2017demy,Risser2017stable} pay attention to iterative optimization. Later, numerous works \cite{Johnson2016perceptual,Ulyanov2016texture,Chen2017stylebank,Zhang2017multi} based on feed-forward network improve both quality and quantity, such as visual effect and computation time. Although improving texture style transfer greatly, these methods transfer only one style via a trained model. A lot of works including AdaIN \cite{Huang2017adain}, WCT \cite{Li2017wct}, Avatar-Net \cite{Sheng2018avatar}, LinearWCT \cite{Li2019lineartransfer}, SANet \cite{park2019arbitrary}, MST\cite{zhang2019multimodal} and recent \cite{liu2021adaattn,hong2021domain,kotovenko2021rethinking,wu2021styleformer,lin2021drafting,wang2020collaborative,xia2020joint,wang2020diversified}, are extended to an arbitrary style transfer. However, these methods are limited to preserving the details of the content image. The problem of the worse content in style transfer has attracted the attention of many scholars. A structure-preserving algorithm \cite{Cheng2019Structure} is introduced to preserve the structure of the content image. ArtFlow \cite{An2021ArtFlow} preserves more details from content image via reversible neural flows. However, their visual qualities are still to be improved. We propose an interest regularization computed by a SuperPoint network \cite{Detone2018superpoint}, which takes the content image as input and outputs the corresponding interest points and descriptors to make the content better.

\subsection{Geometric Style Transfer}
Traditional geometric matching approaches involve detecting and matching hand-crafted interest points, such as SIFT \cite{LoweDavid2004DistinctiveIF}, Shape Context Matching \cite{belongie2002shape} or HOG \cite{Dalal2005HOG}. While these approaches work well for instance-level matching, it is sensitive to appearance variation and noise disturbance. Later, the convolutional neural network has been popular in geometric matching due to its ability to extract powerful and robust features. The current best methods follow the network paradigm proposed by \cite{Rocco2017ConvolutionalNN} which consists of feature extraction, matching layer and regression network, and make various improvements \cite{Rocco2017ConvolutionalNN,Kim2018RecurrentTN,Rocco2018EndtoEndWS,Ham2018ProposalFS,Liu2021warp} based on it. All the above methods act on two RGB images and attempt to estimate a warp field to directly match them. Although performing well between semantically similar images, they fail to deal with objects of different categories with large-scale warping. In the absence of semantic relevance, computing the correlation between two RGB images is unreasonable and defining a match metric is also difficult. DST \cite{Kim2020DeformableST} achieves warping by matching NBB keypoints \cite{aberman2018neural} and estimating thin-plate spline (TPS) \cite{Bookstein1989PrincipalWT} transformation. It is also limited to class-level warping because NBB can only extract key points between similar objects. Some methods are restricted to specialized semantic class such as faces \cite{Yaniv2019TheFO}, caricature \cite{shi2019warpgan} or text \cite{Yang2019textimage_st}. Compared to the above geometric matching method, we achieve large-scale warping even between arbitrary objects of different categories. Overall, different from the above methods, we aim to broaden the style transfer application for product design tasks, and our method obtains amazing industrial product appearances to inspire the designers.

\begin{figure*}[t]
    \centering
    \includegraphics[width=0.97\linewidth]{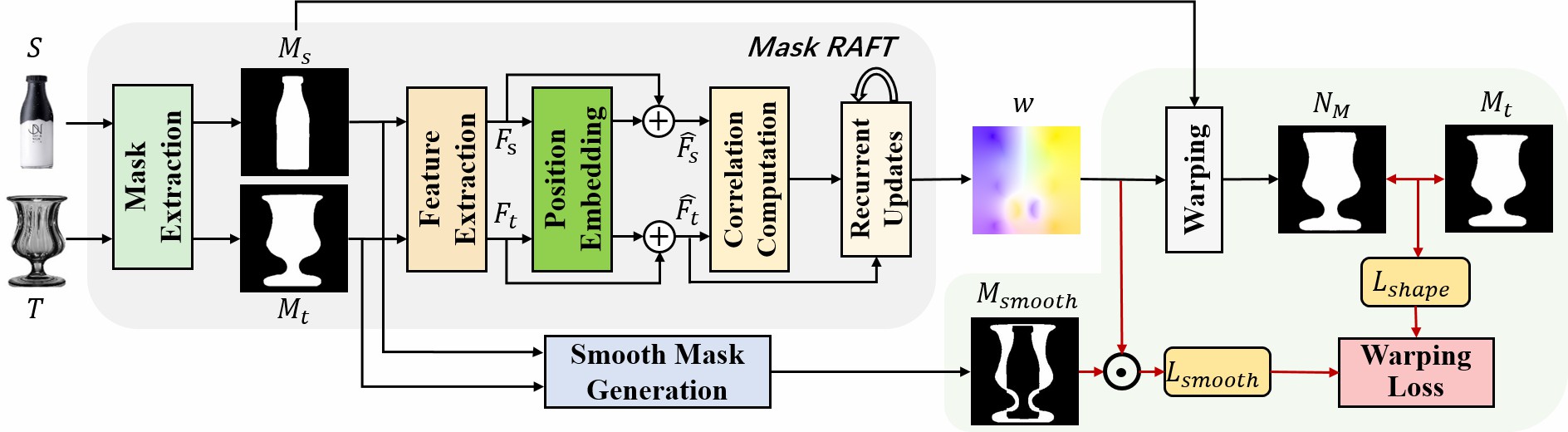}
    \vskip -0.1in
    \caption{Our proposed large-scale geometric warping module including a mask RAFT network and an unsupervised warping loss.}
    \vskip -0.1in
    \label{fig:deform}
\end{figure*}

\section{Industrial Style Transfer}

In this section, we develop an industrial style transfer (InST) framework to create new visual product appearances in Fig. \ref{fig:overall}, consisting of two modules: large-scale geometric warping (LGW) in subsection \ref{sec:deform} and interest-consistency texture transfer (ICTT) in subsection \ref{sec:texture}. We denote a source product (or object) by $S$, a target product $T$, an art reference image by $A$, a new warping product using LGW by $N$, and a final output by $O$.

\subsection{Large-scale Geometric Warping}
\label{sec:deform}

The goal of LGW is to warp a source product $S$ to match the geometric shape of a target product $T$ for a new product generation even if large-scale shape differences and irrelevant semantics. To achieve this goal, we design a neural warping filed between their shape masks inspired by an optical flow method, recurrent all-pairs field transforms (RAFT) \cite{Teed2020RAFTRA}. Specially, Fig.~\ref{fig:deform} shows our LGW module including a mask RAFT and an unsupervised warping loss.

\subsubsection{Mask RAFT Network}
The mask RAFT network can be distilled down to five stages: (1) mask extraction, (2) feature extraction, (3) position embedding, (4) correlation computation, (5) recurrent updates. More details are described as follows.

\textbf{Mask Extraction.} We employ an object segmentation network, denoted as $\mathcal{F}_{m}:\mathbb{R}^{H\times W\times 3}\rightarrow \{0,1\}^{H\times W}$ $\xrightarrow{\text{repeat it 3 times}}\{0,1\}^{H\times W\times 3}$, to extract the masks of the products. Given the products $S$ and $T$ as inputs, their masks are $M_s=\mathcal{F}_{m}(S)$ and $M_t=\mathcal{F}_{m}(T)$, respectively. Here, we use a fixed Resnet50+FPN+PointRend (point-based rendering) network, which has been pre-trained in \cite{kirillov2020pointrend}.

\textbf{Feature Extraction.} Mask features are extracted from the input masks $M_s$ and $M_t$ using a convolutional encoder network, denoted as $\mathcal{F}_{f}:\{0,1\}^{H\times W\times 3}\rightarrow \mathbb{R}^{\frac{H}{8}\times \frac{W}{8}\times D}$, where $D$ is set to $256$. In order to compute the correlation between $M_s$ and $M_t$, the network is similar to the feature encoder network of RAFT \cite{Teed2020RAFTRA}, which consists of $6$ residual blocks, $2$ at $1/2$, $1/4$, and $1/8$ resolutions, respectively. Then, we have the mask multi-scale features, $F_s=\mathcal{F}_{f}(M_s)$, and $F_t=\mathcal{F}_{f}(M_t)$.

\textbf{Position Embedding.}  Due to the lack of color information, there have too many similar or same features between source and target masks, resulting in weak correlation computation and deformation. To avoid such a situation, adjacent position information can improve the deformation field because it updates changes of every pixel (position) of the object product. Thus, we consider position embedding $P$ \cite{vaswani2017attention} of the feature maps $F_{s}$ and $F_{t}$ by using the popular residual operation, and define the new position+feature as:
\begin{align}
\left\{\begin{array}{ll}
\hat{F_s}=F_{s}+P \\
\hat{F_t}=F_{t}+P
   \end{array}
\right.,
\label{math:data_term}
\end{align} where
$P(i,j,k)=
\left\{ 
   \begin{array}{ll}
   \sin\left(\frac{j*\frac{W}{8}+i}{10000^{\frac{k}{D}}}\right), & k\bmod2=0\\
   \cos\left(\frac{j*\frac{W}{8}+i}{10000^{\frac{k-1}{D}}}\right), & k\bmod2=1
   \end{array}
\right.$, $0\leq i\leq \frac{H}{8}$, $0\leq j\leq \frac{W}{8}$, and $0\leq k\leq D=256$.

\textbf{Correlation Computation and Recurrent Updates.} Here, we follow the computing visual similarity and iterative updates of RAFT \cite{Teed2020RAFTRA} to compute the multi-scale correlations and recurrently update the warping field. In this paper, these two steps are denoted as $\mathcal{F}_{cr}:(\mathbb{R}^{H/8\times W/8\times D},\mathbb{R}^{H/8\times W/8\times D})\rightarrow \mathbb{R}^{H\times W\times 2}$. Overall, our mask RAFT network is described as: $\omega=\{\omega_r\}_{r=1}^R$
\begin{align}
\label{math:maskwarp}
=\mathcal{F}_{cr}\left(\mathcal{F}_{f}\left(\mathcal{F}_{m}(S)\right)+P,\ \mathcal{F}_{f}\left(\mathcal{F}_{m}(N)\right)+P\right),
\end{align} where $R$ is the number of iterations, and we set $R=3$ in our implementation.

\subsubsection{Unsupervised Warping Loss}
The mask RAFT network is trained in an unsupervised setting by constructing a shape-consistency loss, and a smoothness regularization.

\textbf{Shape-consistency loss.} Based on the warping filed estimation $\omega$ in Eq.~\eqref{math:maskwarp}, we get the warped source masks $\{\omega_r(M_s)\}_{r=1}^R$ by the differential bilinear sampling mentioned in spatial transformer \cite{Jaderberg2015SpatialTN}. Given the target mask $M_t$, this $\ell_1$ loss is defined as
\begin{align}
\label{math:shapeloss}
L_{shape} = \sum\nolimits_{r=1}^{R}\alpha_{r}\|\omega_{r}(M_{s})-M_{t}\|_1,
\end{align}
where $\alpha_r$ is used to balance the deformation degree.

\begin{figure}[t]
    \centering
    \includegraphics[width=0.97\linewidth]{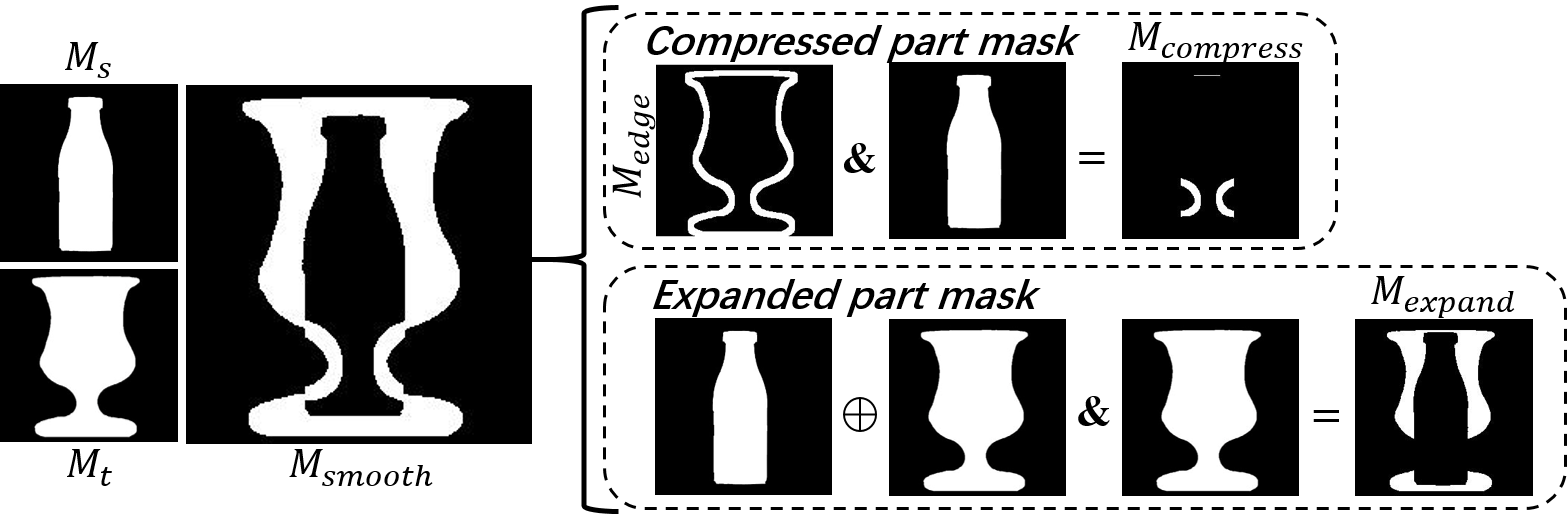}
    \vskip -0.1in
    \caption{Given two shapes, we design different smoothness masks for the two parts of compression (upper right) and expansion (lower right). The smoothness mask in the middle is what we use in the smoothness regularization.}
    \label{fig:smooth_mask}
    \vskip -0.1in
\end{figure}

\textbf{Smoothness regularization.} To avoid the chaotic deformation, it needs to further restrict the sampling direction of the warping field for maximally holding the content detail of the source object. Specially, we design a smoothness mask, shown in Fig.~\ref{fig:smooth_mask}, and its generated formula is expressed as
\begin{align}
\label{math:smoothmask}
M_{smooth}&=M_{compress}|M_{expand} \nonumber \\
&=(M_{edge}\&M_s)|(M_s\oplus{M_t}\&M_t),
\end{align}where $|$, $\&$ and $\oplus$ denote logical disjunction, conjunction, and XOR, $M_{edge}$ represents edges of the target object product. $M_{edge}$ is computed by convolution operation with all one kernel, $M_{edge}=Cov(M_t,ker)$, where $ker=[1]^{k\times k\times 3}$, $k$ is a predefined kernel size, and we set $k=9$.
(More details are provided in supplementary materials.)
Since $M_{smooth}\in \{0,1\}^{H\times W\times 3}$ has same mask maps in three channels, one channel is denoted by $\mathbf{M}\in \{0,1\}^{H\times W}$.
Given the warp field estimations  $\{\omega_r(M_s)\}_{r=1}^R$, the $\ell_2$ regularization on $\mathbf{M}$ is defined as
\begin{align}
\label{math:smoothness}
L_{smooth} = \sum\nolimits_{r=1}^{R}\beta_{r}L_{smooth}(\omega_r,\mathbf{M}),
\end{align}
where $\beta_r$ denotes the degree of content retention of different warp fields, and $L_{smooth}(\omega_r,\mathbf{M}) =$
\begin{align}
&\frac{1}{\sum\nolimits_{i,j}\mathbf{M}_{ij}}\sum\nolimits_{i,j}\mathbf{M}_{ij}\left(\Vert \omega_{r}^{i+1,j}-\omega_{r}^{i,j}\Vert_2 +\Vert \omega_{r}^{i,j+1}-\omega_{r}^{i,j} \Vert_2 \right. \nonumber \\
& \ \ \ \ \ \ \ \ \ \  \left.+\Vert \omega_{r}^{i+1,j+1}-\omega_{r}^{i,j} \Vert_2+\Vert \omega_{r}^{i+1,j-1}-\omega_{r}^{i,j} \Vert_2\right).
\end{align}The above term is a first-order smoothness on the warp field $\omega$ by constraining the displacement of horizontal, vertical, and diagonal neighborhoods around coordinate $(i,j)$. It drives the texture content of the source object to be close to its neighborhoods after the deformation. By combining $L_{shape}$ with $L_{smooth}$, the warping loss is described as
\begin{equation}
\label{math:warp_overall}
	L_{overall} = L_{shape} + \gamma L_{smooth},
\end{equation}
where $\gamma=1$ controls the importance of each term.

\subsection{Interest-Consistency Texture Transfer}
\label{sec:texture}
\begin{figure}[t]
    \centering
    \includegraphics[width=0.95\linewidth]{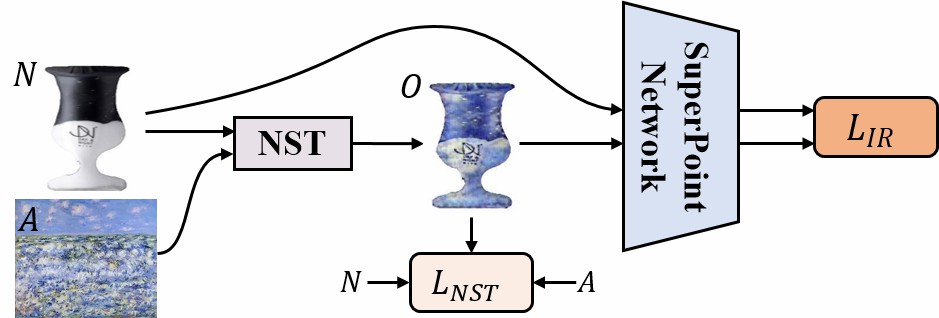}
    \vskip -0.1in
    \caption{Interest-consistency texture transfer. It consists of a NST method for artistic style transformation and a SuperPoint network for content preservation by interest point constraints. }
    \vskip -0.2in
    \label{fig:style}
\end{figure}

After generating the new product $N$ by LGW, the goal of ICTT is to create a stylized product appearance $O$ with important content details of $N$ by transferring the art style of reference image $A$ to $N$ using neural style transfer (NST) methods. To achieve this goal, we introduce an interest regularization (IR) term in Fig.~\ref{fig:style} to maintain similarity between interesting contents of $O$ and $N$ based on the SuperPoint network \cite{Detone2018superpoint} as it can effectively compute interest point locations and their associated descriptors.

\textbf{NST} is often to train an image transformation network $\mathcal{F}$ by minimizing a NST loss, denoted as $L_{\text{NST}}$, including both content and texture style losses. In this work, we consider two popular algorithms, AdaIN \cite{Huang2017adain} and LinearWCT \cite{Li2019lineartransfer}, and one most relevant method, ArtFlow \cite{An2021ArtFlow}.

\textbf{IR} is to control perceptual differences between $N$ and $O$ by the SuperPoint network, denoted as $\mathcal{S}(\cdot)$, outputting an interest point head with $H\times W$ size and $65$ channels, $\mathcal{P}\in \mathbb{R}^{H\times W\times 65}$, and descriptor head with $H\times W$ size and $256$ channels, $\mathcal{D}\in R^{H\times W\times 256}$. Then we have $(\mathcal{P}_{N}, \mathcal{D}_{N})=\mathcal{S}(N)$, and $(\mathcal{P}_{O}, \mathcal{D}_{O})=\mathcal{S}(O)$. IR is defined as follows:
\begin{align}
\label{eq:irloss}
L_{\text{IR}}= L_{\mathcal{P}}(\mathcal{P}_{N}, \mathcal{P}_{O}) + \lambda L_{\mathcal{D}}(\mathcal{D}_{N}, \mathcal{D}_{O}),
\end{align}
where $\lambda=0.00005$. $L_{\mathcal{P}}$ is a square of $\ell_2$ norm, that is,
\begin{align}
L_{\mathcal{P}}(\mathcal{P}_{N}, \mathcal{P}_{O})=\frac{1}{HW}\sum_{h=1}^H\sum_{w=1}^W\|\mathbf{p}_{hw}^N-\mathbf{p}_{hw}^O\|_2^2,
\end{align} where
$\mathbf{p}_{hw}^N$ and $\mathbf{p}_{hw}^O$ are $65$-dimensional vectors belonged to $\mathcal{P}_{N}$ and $ \mathcal{P}_{O}$, respectively. $L_{\mathcal{D}}$ is a hinge loss \cite{Detone2018superpoint} with positive margin $m_p=1$ and negative margin $m_n=0.2$, that is, $L_{\mathcal{D}}(\mathcal{D}_{N}, \mathcal{D}_{O})=$
\begin{align}
\frac{1}{(HW)^2}\sum_{h=1}^H\sum_{w=1}^W\sum_{i=1}^H\sum_{j=1}^W l_d(\mathbf{d}_{hw}^N,\mathbf{d}_{ij}^O;g_{hwij}),
\end{align} where $l_d(\mathbf{d}^N,\mathbf{d}^O;g)=\beta g \max(0,m_p-(\mathbf{d}^N)^T\mathbf{d}^O)+(1-g)\max(0,(\mathbf{d}^N)^T\mathbf{d}^O-m_n)$, $g_{hwij}$ is a homography-induced correspondence between $(h,w)$ and $(i,j)$ cells, $g_{hwij}=
\left\{
   \begin{array}{ll}
   1, & \hbox{if $\|\widehat{\mathcal{H}\mathbf{h}_{hw}^N}-\mathbf{h}_{ij}^O\|\leq 8$,}\\
   0, & \hbox{otherwise.}
   \end{array}
\right.$, $\mathbf{h}_{hw}^N$ denotes the location of the center pixel in the $(h,w)$ cell, and $\widehat{\mathcal{H}\mathbf{h}_{hw}^N}$ denotes multiplying the cell location $\mathbf{h}_{hw}^N$ by the homography $\mathcal{H}$ and dividing by the last coordinate. By combining $L_{\text{NST}}$ with $L_{\text{IR}}$, the ICTT loss is described as
\begin{align}
\label{eq:icttloss}
L_{\text{ICTT}}=L_{\text{NST}}+\mu L_{\text{IR}},
\end{align} where $\mu=1$ controls the balance between NST and IR.

\begin{figure*}[ht]
    \centering
    \includegraphics[width=0.90\linewidth]{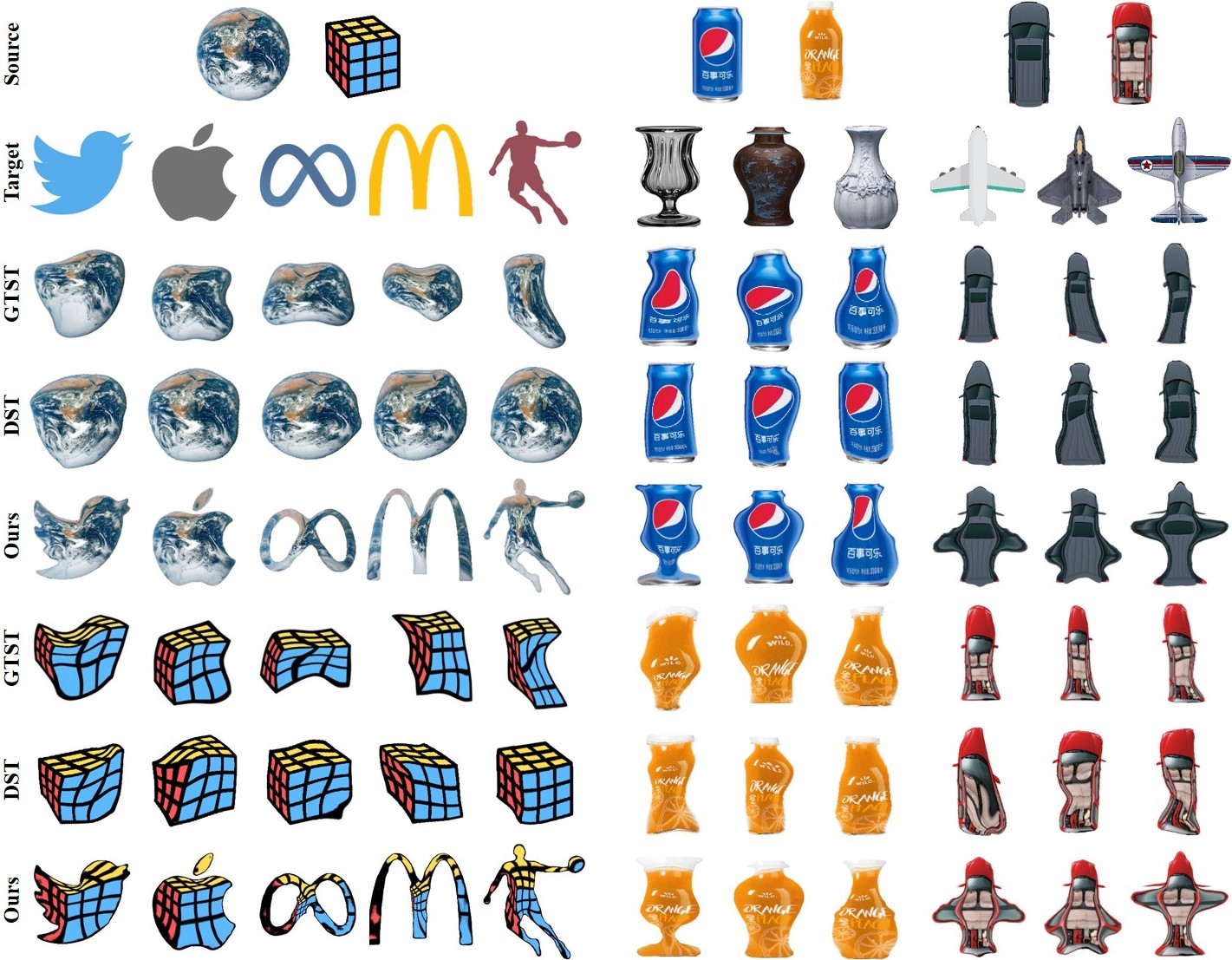}
    \vskip -0.1in
    \caption{Visual product design results using the geometric style transfer methods, \eg, DST \cite{Kim2020DeformableST}, GTST \cite{Liu2021warp} and our InST. Compared to DST and GTST, our intermediate results between cars and aircraft have more reference values for the product designers as they are similar to the top view of the products (\eg, Terrafugia and AeroMobil-4.0\protect\footnotemark[1])
    }
     \vskip -0.05in
    \label{fig:geometricproduct}
\end{figure*}

\begin{figure*}[ht]
\centering
\includegraphics[width=0.90\linewidth]{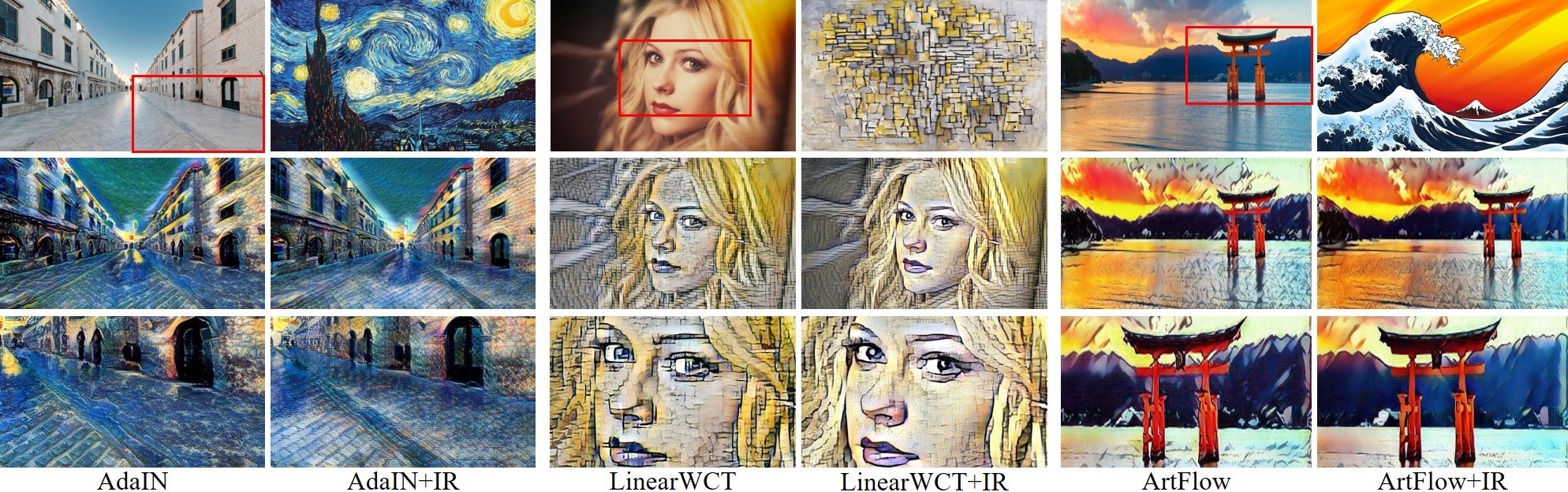}
\vskip -0.1in
\caption{Content preservation results using the texture style transfer methods, \eg, AdaIN \cite{Huang2017adain}, LinearWCT \cite{Li2019lineartransfer} and ArtFlow \cite{An2021ArtFlow}. }
\vskip -0.15in
\label{fig:texturetransfer}
\end{figure*}

\begin{figure*}[ht]
    \centering
    \includegraphics[width=0.90\linewidth]{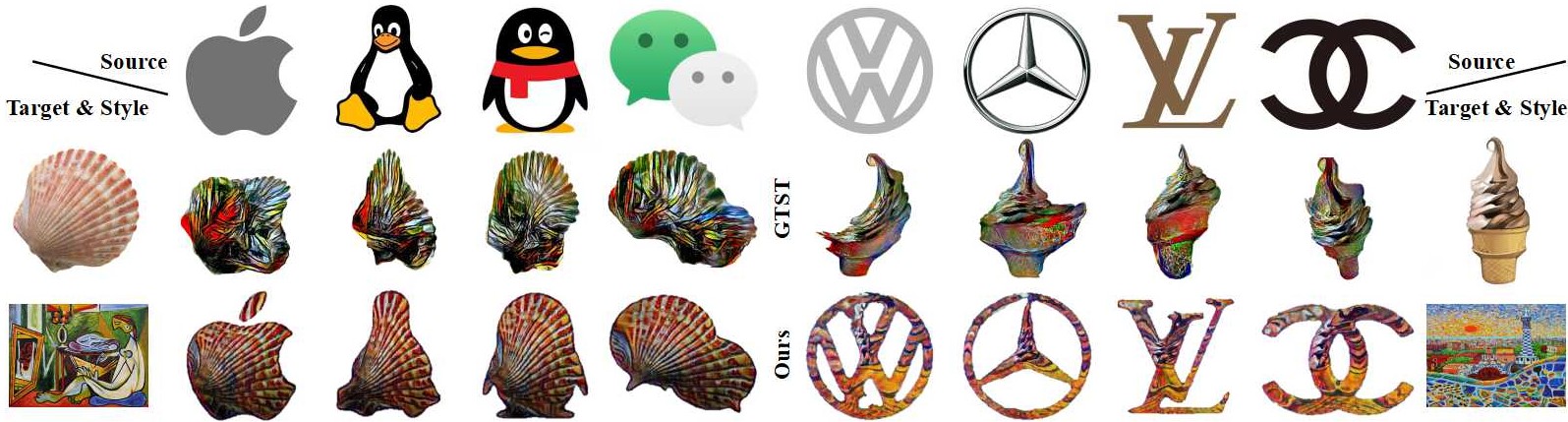}
    \vskip -0.1in
    \caption{Visual logo design results using the geometric and texture style transfer methods, \eg, GTST \cite{Liu2021warp} and our InST.}
    \vskip -0.05in
    \label{fig:stylelogo}
\end{figure*}
\begin{figure*}[ht]
    \centering
    \includegraphics[width=0.90\linewidth]{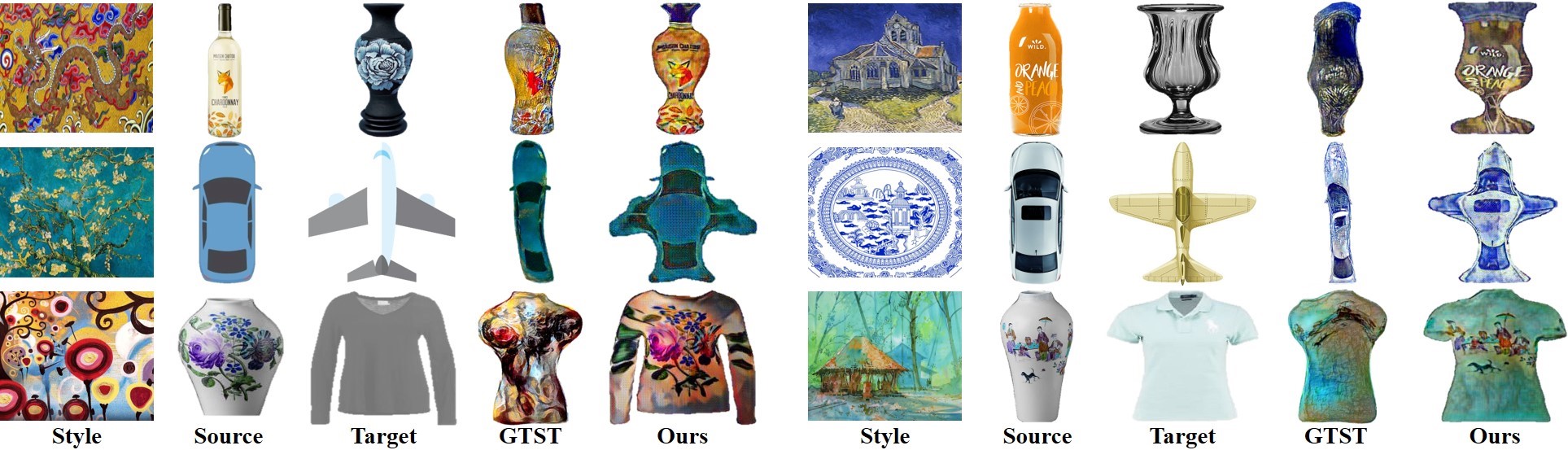}
    \vskip -0.1in
    \caption{Visual product design results using the geometric and texture style transfer methods, \eg, GTST \cite{Liu2021warp} and our InST.}
    \vskip -0.15in
    \label{fig:styleindus}
\end{figure*}

\section{Experiments}
In this section, we conduct extensive experiments to evaluate the visual product design ability of our InST approach, \eg, company logos, bottles, porcelain fashions, and flying cars. More comparisons of product design are available in supplementary materials.


\subsection{Experimental Settings}


\noindent{\bf Dataset.} consists of source and target products (or objects) and art style images. Following \cite{Wu2021DerenderingTW}, the source products are selected from the Metropolitan Museum of Art Collection via the open-access API \cite{2020museum}, and their segmentation masks are obtained by using the PointRend \cite{kirillov2020pointrend}. We use the clothes collected from the Zalando dataset \cite{Jetchev2017Zalando} as the target products and get their segmentation masks by using the VITON \cite{Han2018VITONAI}. The art style images are the WikiArt dataset \cite{http2016wikiart}. In addition, the MS-COCO dataset \cite{Lin2014coco} is also considered as the content images for training the network in the ICTT module. The input images are resized to 512$ \times $512. Each image is randomly cropped to 256$ \times $256 for training.

\noindent{\bf Training.} Since our model includes the LGW and ICTT modules, our training schedule is decomposed into three steps. First, the source and target products are used to train the warping network of LGW. The hyper-parameters are set to $\{\alpha_r\}_{r=1}^3=\{0.1,0.2,1\}$ in Eq. \eqref{math:shapeloss}, $\{\beta_r\}_{r=1}^3=\{0.1,0.05,0.01\}$ in Eq. \eqref{math:smoothness}, and $\gamma=1$ in Eq. \eqref{math:warp_overall}. Second, the art style images and MS-COCO as the content images are used to train the art transfer network of ICTT. The hyper-parameters are set to $\lambda = 0.00005$ in Eq.~\eqref{eq:irloss}, and $\mu=1$ in Eq.~\eqref{eq:icttloss}. Third, we jointly optimize the both the warping and art transfer networks using the collected dataset. In our experiment, We train these three steps for 50k/60k/10k iterations with a batch size 16/2/2 and the Adam \cite{kingma2015adam} optimizer with a learning rate of 0.001/initial 0.0001 with decay of 0.00001/0.0001. Training takes roughly 10/12/8 hours on a single GTX 2080Ti GPU.

\subsection{Main Results}
To demonstrate that the proposed InST has a geometric and texture transfer ability to create a new product with wonderful visual appearances, we compare it with two recent geometric transfer methods, \eg, DST \cite{Kim2020DeformableST} and GTST \cite{Liu2021warp}, and three texture transfer methods, \eg, AdaIN \cite{Huang2017adain}, LinearWCT \cite{Li2019lineartransfer}, and ArtFlow (content preservation) \cite{An2021ArtFlow}.

\noindent{\bf Visual comparisons.}
We showcase new visual products qualitatively from three aspects: (i) geometric warping, (ii) texture transfer, (iii) their combination.

\noindent{\emph{Geometric warping.}} Fig.~\ref{fig:geometricproduct} shows the new product design results by the geometric style transfer algorithms. For example, the round earth and Rubik's cube are transferred into the logos of Twitter, Apple, Meta, McDonald's, and Jordan, respectively. Compared to the geometric methods, \eg, DST and GTST, our LGW module can better match the geometric shape of the target and better maintain the texture content of the source. The reasons for their failures are that DST and GTST only have little semantic relationships between two objects by using the corresponding key points \cite{Kim2020DeformableST} and learning a small-scale warping field \cite{Liu2021warp}, resulting in a worse result when facing large-scale geometric shapes. In contrast, we design a smooth mask warping field to adapt for large-scale warping in the visual product design.

\footnotetext[1]{\url{https://www.beautifullife.info/automotive-design/10-real-flying-cars/}}

\noindent{\emph{Texture transfer.}} Fig.~\ref{fig:texturetransfer} shows the content preservation of the texture style transfer algorithms, \eg, AdaIN, LinearWCT and ArtFlow. We can observe that our IR regularization can improve all the algorithms to preserve more content details since it perceives interest points to be similar. This is very different from ArtFlow as it considers reversible neural flows and unbiased feature transfer.

\begin{table}
\linespread{1}
\centering
\setlength{\tabcolsep}{7pt}
\renewcommand{\arraystretch}{1.1}
\resizebox{0.65\linewidth}{!}{
\begin{tabular}{c|ccc}
\hline
Method &DST \cite{Kim2020DeformableST} &GTST \cite{Liu2021warp} &our LGW \\
\hline
mIoU $ \uparrow $ &0.6000 &0.7285 &{\bf 0.9284} \\
\hline
\end{tabular}}
\vskip -0.15in
\caption{Quantitative evaluations of geometric warping methods.}
\vskip -0.1in
\label{tab:miou}
\end{table}

\begin{table}
\linespread{0.8}
\centering
\setlength{\tabcolsep}{2pt}
\renewcommand{\arraystretch}{1.1}
\resizebox{1\linewidth}{!}{
\begin{tabular}{c|cccccc}
\hline
		Method &AdaIN \cite{Huang2017adain}  &AdaIN+IR &LinWCT \cite{Li2019lineartransfer} &LinWCT+IR &ArtFlow \cite{An2021ArtFlow} &ArtFlow+IR \\
		\hline
		SSIM $ \uparrow $ &0.3424 &{\bf0.3886} &0.4612 &{\bf 0.4932} &0.5042 &{\bf 0.5643}       \\
		Time(s) $ \downarrow $ &0.054 &0.054 &0.419 &{\bf 0.416} &{\bf 0.138} &0.140\\
\hline
\end{tabular}}
\vskip -0.1in
\caption{Quantitative evaluations of stylization methods.}
\vskip -0.2in
\label{tab:ssim}
\end{table}

\noindent{\emph{Geometric\&texture transfer.}} We evaluate the overall product design with beautiful appearances based on the combination of geometric and texture style transfer against the state-of-the-art GTST \cite{Liu2021warp}. Fig.~\ref{fig:first} shows that our InST method is to create wonderful product appearances, such as the snail logo of Apple and Twitter. In addition, Figs.~\ref{fig:stylelogo} and \ref{fig:styleindus} also show more product design results. Compared to GTST, our method can provide larger-scale warping and retain more details of the source object (or product).

\noindent{\bf Quantitative comparisons.}
In addition to the above visual comparisons, we provide two quantitative comparisons for the LGW and IR modules. First, we evaluate the geometric warping performance with mean intersection-over intersection (mIoU), which is the popular metric for semantic segmentation \cite{long2015fully}. In Table~\ref{tab:miou}, we see that LGW has higher mIoU scores than DST and GTST. This means that the warping product better matches the target's geometry. Second, similar to \cite{An2021ArtFlow}, the Structural Similarity Index (SSIM) between the content and stylized images is considered as a metric to measure the performance of the detail preservation. Table~\ref{tab:ssim} reports that these methods using our IR term have higher SSIM scores and can preserve more detailed information without additional test times.

\begin{table}
\linespread{0.8}
\centering
\setlength{\tabcolsep}{2pt}
\renewcommand{\arraystretch}{1.1}
\resizebox{1\linewidth}{!}{
\begin{tabular}{c|ccc|cc|cc}
\hline
\multirow{2}{*}{Method} & \multicolumn{3}{c|}{Geometry} & \multicolumn{2}{c|}{Texture} &\multicolumn{2}{c}{Combination} \\
\cline{2-8}
  & DST \cite{Kim2020DeformableST} & GTST \cite{Liu2021warp} & Ours & TextMethod & Ours & GTST \cite{Liu2021warp} & Ours         \\
\hline
Votes$\uparrow$ &37 &60 &{\bf 1043} &337 &{\bf 763} &134 &{\bf 1006} \\
\hline
\end{tabular}}
\vskip -0.1in
\caption{Quantitative evaluations of user study. TextMethod denotes a set of AdaIN \cite{Huang2017adain}, LinWCT \cite{Li2019lineartransfer} and ArtFlow \cite{An2021ArtFlow}. }
\vskip -0.2in
\label{tab:votes}
\end{table}

\noindent{\bf User Study.} We conduct a user study to evaluate the effect of the proposed InST algorithm against the existing methods. We divide the evaluation into three groups from the perspectives of geometric warping, content maintenance, and their combination, and each group includes ten options. In total, we collected 3420 votes from 114 users, and each group gets 1140 votes. Table~\ref{tab:votes} reports the results of the specific votes. Given the source and target products, 91.5\% of users reported that our LGW network better matches the target's geometry, compared to only 5.3\% for GTST \cite{Liu2021warp} and 3.2\% for DST \cite{Kim2020DeformableST}. In the content maintenance evaluation, 66.9\% of users thought our ICTT module maintains more content details than the corresponding texture style transfer methods \cite{Huang2017adain,Li2019lineartransfer,An2021ArtFlow}. Finally, when evaluating the overall effect from the above two aspects, our proposed algorithm accounted for 88.2\% of the 1140 votes compared to 11.8\% for GTST \cite{Liu2021warp}. Overall, our results were favorite among all aspects and evaluated methods.



\subsection{Ablation Study}
\begin{figure}[ht]
    \vskip -0.05in
    \centering
    \includegraphics[width=0.97\linewidth]{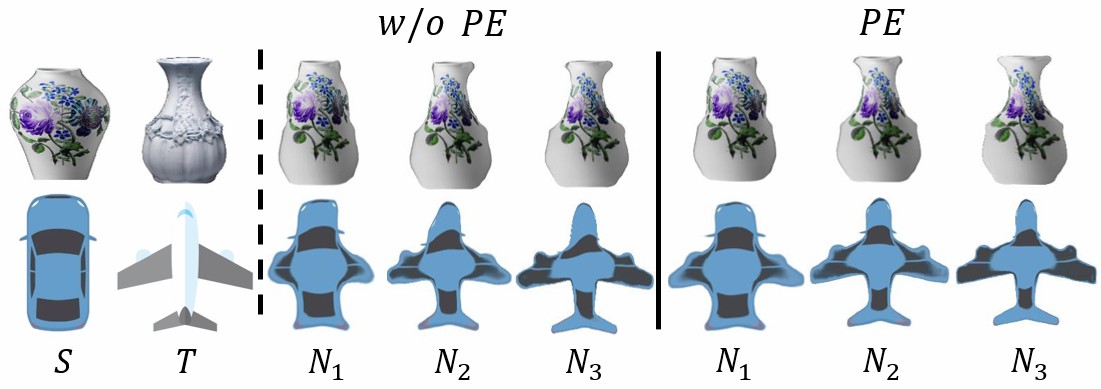}
    \vskip -0.1in
    \caption{Ablation study on position embedding.}
    \label{fig:PE_ablation}
    \vskip -0.1in
\end{figure}

Since the comparative experiments on LWG and IR of ICTT have been available in the above subsection, we perform an ablation experiment on position embedding of the mask RAFT network in LWG. We test the importance of the position embedding by training an LGW module without this component. Fig.~\ref{fig:PE_ablation} shows the comparison results with three recurrent updates. The position embedding achieves better performance because such an operation enhances the correlation of adjacent positions.


\begin{figure}[t]
    \centering
    \includegraphics[width=0.97\linewidth]{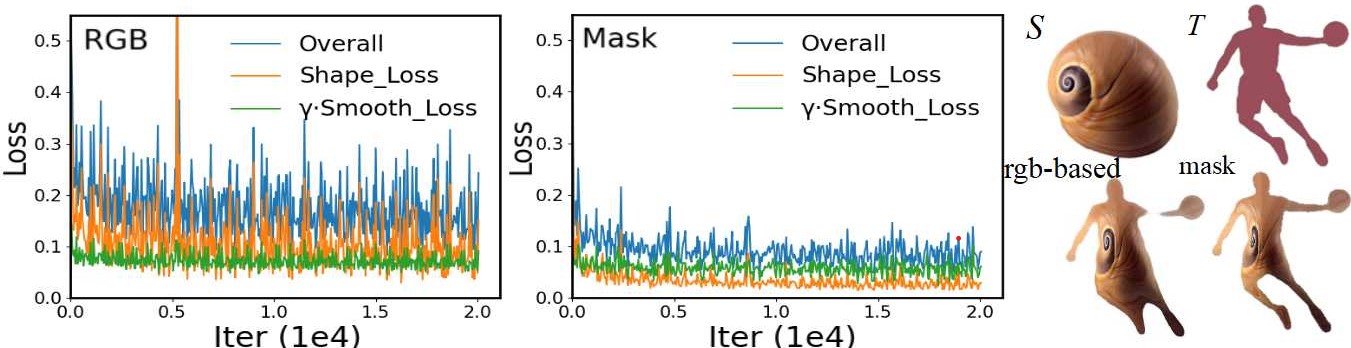}
    \vskip -0.1in
    \caption{Losses of our LGW module with RGB and mask inputs.}
    \label{fig:loss_ablation}
    \vskip -0.2in
\end{figure}

\section{Discussion}
In this section, we discuss three problems to better understand our mask RAFT and the limitations of our InST method. In addition, the potential applications are available in supplementary materials.

\emph{Why are RAFT \cite{Teed2020RAFTRA} suitable for the geometric warping task?} There are three reasons for explanations. 1) Optical-flow estimation is widely applied to estimate a warping between two moving geometries of objects in consecutive video frames by learning a warp field \cite{long2015fully, Ranjan2017OpticalFE,sun2018pwc, Hur2019IterativeRR, Teed2020RAFTRA}. 2) Similar to the optical-flow estimation, a semantic transformer method \cite{Kim2018RecurrentTN} has been used to train a geometric warping filed between similar objects, named GTST \cite{Liu2021warp}, which is better than DST \cite{Kim2020DeformableST}. 3) RAFT \cite{Teed2020RAFTRA} is state-of-the-art as it got the best paper award of ECCV 2020.

\emph{Why do we design a mask warping field?} One reason is that it is difficult or impossible to directly warp RGB pixels of one object to match another when they are semantically irrelevant or have a big difference between their shapes, such as snail and Twitter logo. Another reason is that the difference between the two masks is lower than textural RGB images, leading to easier optimization. We train our LGW module with RGB images and their masks inputs and show the loss curves in Fig.~\ref{fig:loss_ablation}. It is clear that using masks inputs has much lower loss and faster convergence than RGB. For further comparison, we also show their visualization results separately in Fig.~\ref{fig:loss_ablation}, and obviously, mask RAFT has better deformation than RGB-based RAFT.

\emph{What differences between RAFT and mask RAFT?}
Compared to RAFT \cite{Teed2020RAFTRA}, our mask RAFT has the following four differences. Firstly, we design an unsupervised loss and a mask smoothness to learn a large-scale warping field, while RAFT explores a small-scale light flow field in a supervised setting. Secondly, before RAFT, we introduce a mask extraction stage to obtain object (or product) mask from its RGB image. Thirdly, we present a position embedding for the feature extraction behind enhancing the correlation of adjacent positions. Fourthly, we use the feature $\hat{F_t}$ of the target instead of the feature extraction using another network. Overall, our mask RAFT can better warp large-scale geometric shapes.

\begin{figure}[h]
    \centering
    \includegraphics[width=1\linewidth]{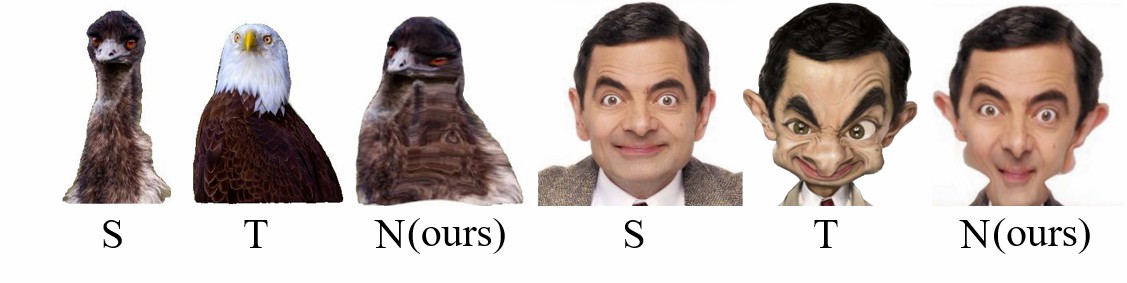}
    \vskip -0.15in
    \caption{Limitation: The in-principle limit is semantic corresponding between similar objects.}
    \vskip -0.2in
    \label{fig:bad_result}
\end{figure}

\emph{Limitations.} Here, we discuss the limitations of geometric warping. Because our purpose is to realize a large-scale warping field between products (or objects), which has a little semantic correspondence, we do not rely on semantic information to guide the warping field. When input pairs share semantic attributes, our method may generate counterintuitive results. For example, in Fig.~\ref{fig:bad_result}, our LGW method tries to match the shapes regardless of internal semantic alignment, such as aligning eyes with eyes.

\section{Conclusion}

In this paper, we proposed an industrial style transfer method for visual product design tasks. Our method constructed a geometric transformation field to create a new product and further learned a style transformation network to transfer an art style of the reference image to the new product. It is worth mentioning that our method warped a source product to imitate the geometric shape of the target product even if they are not semantically relevant. Extensive experiments demonstrated that our method outperforms the state-of-the-art style transfer algorithms, particularly the challenging large-scale geometric shapes. We also applied the style transfer pipeline into some product design tasks, \eg, amazing logos, beautiful bottles, flying cars and porcelain fashions. Hopefully, our work can open an avenue to assist or inspire designers to design new industrial products by using style transfer techniques.


{\small
\bibliographystyle{ieee_fullname}
\bibliography{egbib}
}

\clearpage
\setcounter{table}{0}
\setcounter{figure}{0}
\setcounter{equation}{0}
\setcounter{section}{18}
\renewcommand\thesection{\Alph{section}}
\renewcommand{\thetable}{S\arabic{table}}
\renewcommand{\thefigure}{S\arabic{figure}}
\renewcommand{\theequation}{S\arabic{equation}}
\section{Supplementary Material}
This supplementary material consists of the following five parts: an explanation of the smoothness mask generation (subsection \ref{sec:smoothness_mask}), intermediate results of different deformation degrees (subsection \ref{sec:intermediate_results}), potential applications (subsection \ref{sec:application}), a description of our evaluation interface for the user study (subsection \ref{sec:user_study}), and more comparison results (subsection \ref{sec:more_results})

\subsection{Smoothness Mask Generation}
\label{sec:smoothness_mask}

\begin{figure}[h]
    \centering
    \includegraphics[width=0.9\linewidth]{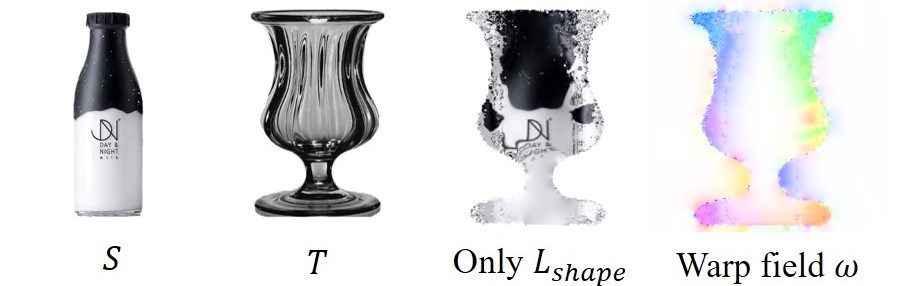}
    \caption{The shape-consistency only ensures the shape matching between masks and cannot further restrict the sampling direction, which may cause chaotic deformation on the image domain.}
\label{fig:only_Lshape}
\end{figure}

Relying only on the shape-consistency loss is not enough because there exist many distinct warp fields that can make the loss minimized. Due to the lack of guidance on sampling direction, although it can achieve shape matching between masks, the post-deformation result on the image domain may be very random and chaotic, as shown in Fig.~\ref{fig:only_Lshape}. Therefore, we must further restrict the sampling direction of the warping field to retain the content details of the source object, to the greatest extent. In this work, we abide by the deformation rule of an edge-to-edge sampling, which can make full use of the source content, and propose a mask smoothness regularization based on it.

\begin{figure}[h]
    \centering
    \includegraphics[width=0.9\linewidth]{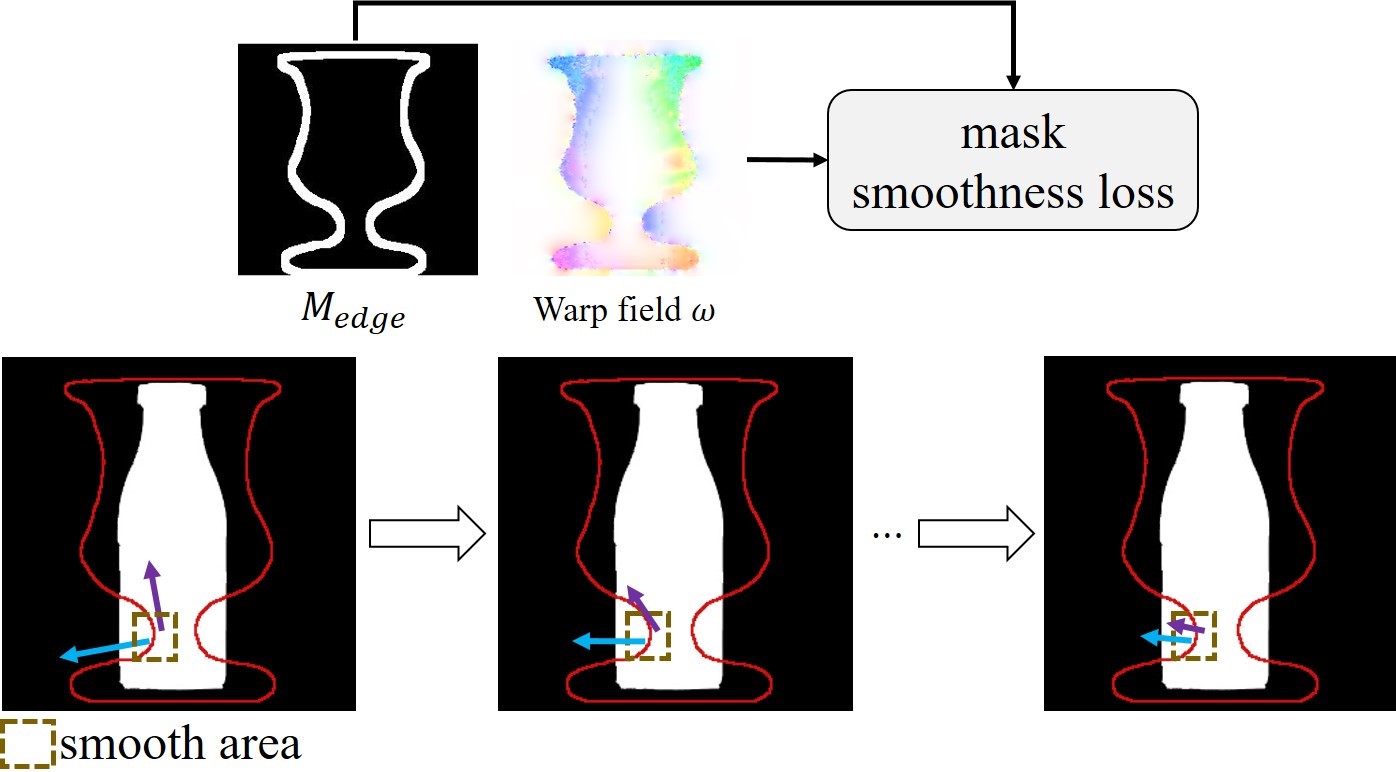}
    \caption{By adding a smoothness constraint on both sides of the target edge($M_{edge}$), the target edge region (take brown square area as example) tend to sample from the source edge.}
    \label{fig:edge_smooth}
\end{figure}

To achieve the goal of the edge-to-edge sampling, under the shape-consistency loss, we add a smoothness constraint on both sides of the target edge ($M_{edge}$) to the warp field, as illustrated in Fig.~\ref{fig:edge_smooth}. Firstly, the displacement on both sides of the target edge should be as close as possible. Secondly, the shape and background areas of the target will sample from the shape (purple arrow) and background (blue arrow) areas of the source, respectively. Combined with the two constraints described above, the target edge region will tend to sample from the source edge.

More specially, we divide the deformation into two parts, a compressed part (like the handle part of the cup in Fig.~\ref{fig:edge_smooth}) and an expanded part (like the body and bottom part of the cup). When there is no constraint on the shape-consistency loss, Fig.~\ref{fig:only_Lshape} shows that the compressed part looks like an edge truncation and the extended part randomly samples from the source shape area. To overcome the truncation problem, we add the edge smoothness mask into the region for imposing the edge-to-edge sampling in the compressed part. To avoid the blur problem, we use the smoothness constraint in the extended part. As a result, we explore the smoothness mask in the form shown in Fig.~\ref{fig:smooth_mask}.

\subsection{Intermediate Results of LGW}
\label{sec:intermediate_results}

Since the LGW module generates a prediction sequence, we expect to reflect the deformation process. Because the latter warp field produces a larger scale warping, we use increasing weights to balance the shape-consistency loss sequence between the warped source mask and the target mask. As the latter warp field needs to relax smoothness requirements more than the former, we employ decreasing weights to balance the smoothness loss sequence. They are described as
\begin{align}
\label{smath:shapeloss}
L_{shape} &= \sum_{r=1}^{R}\alpha_{r}\|\omega_{r}(M_{s})-M_{t}\|_1, \\
L_{smooth} &= \sum_{r=1}^{R}\beta_{r}L_{smooth}(\omega_r,\mathbf{M}),
\end{align}
where $R$ is the number of iterations, and we set $R=3$ in our implementation. The increasing weights
$\{\alpha_r\}_{r=1}^3$ are set to $\{0.1,0.2,1\}$, and the decreasing weights $\{\beta_r\}_{r=1}^3$ are set to $\{0.1,0.05,0.01\}$.

\begin{figure}[ht]
    \centering
    \includegraphics[width=0.8\linewidth]{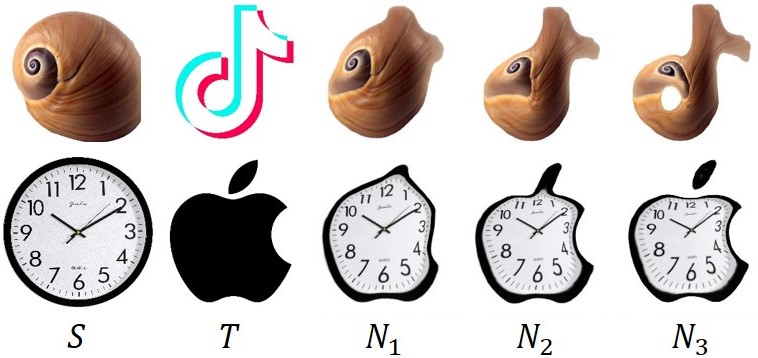}
    \vskip -0.1in
    \caption{Intermediate results $\{N_r\}_{r=1}^3$ with three warping fields $\{\omega_r\}_{r=1}^3$. }
    \label{fig:intermediate_results}
    \vskip -0.2in
\end{figure}

Once trained, the LGW network can be used to generate warp fields iteratively which become increasingly accurate with respect to the geometric shape of the target, as shown in Fig.~\ref{fig:intermediate_results}.

\subsection{Potential Applications}
\label{sec:application}

\begin{figure}[h]
    \centering
    \includegraphics[width=0.9\linewidth]{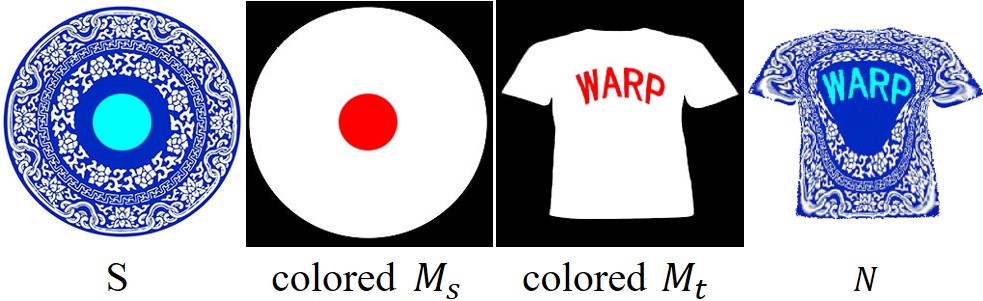}
    \caption{Restricting the sampling direction by using colored masks. Here, we expect the specific \emph{WARP} area in the cloth to be warped from the cyan circle in the plate.}
  \label{fig:colored_mask}
\end{figure}

In addition to the smoothness regularization applied to the warp field, we can also restrict the sampling direction by dividing different corresponding areas into many masks. As shown in Fig.~\ref{fig:colored_mask}, we expect the specific \emph{WARP} area in the colored $M_t$ to be warped from the cyan circle in $S$. We achieve it by iterating a warp field based on $L_{shape}$ and $L_{smooth}$ between $M_s$ and $M_t$. $N$ is the finally warped result. So, depending on users' preferences, they can customize the various artistic product, for example, the design of clothing with a logo in Fig.~\ref{fig:colored_mask}.

\subsection{User Study}
\label{sec:user_study}
\begin{figure}[ht]
    \centering
    \includegraphics[width=0.9\linewidth]{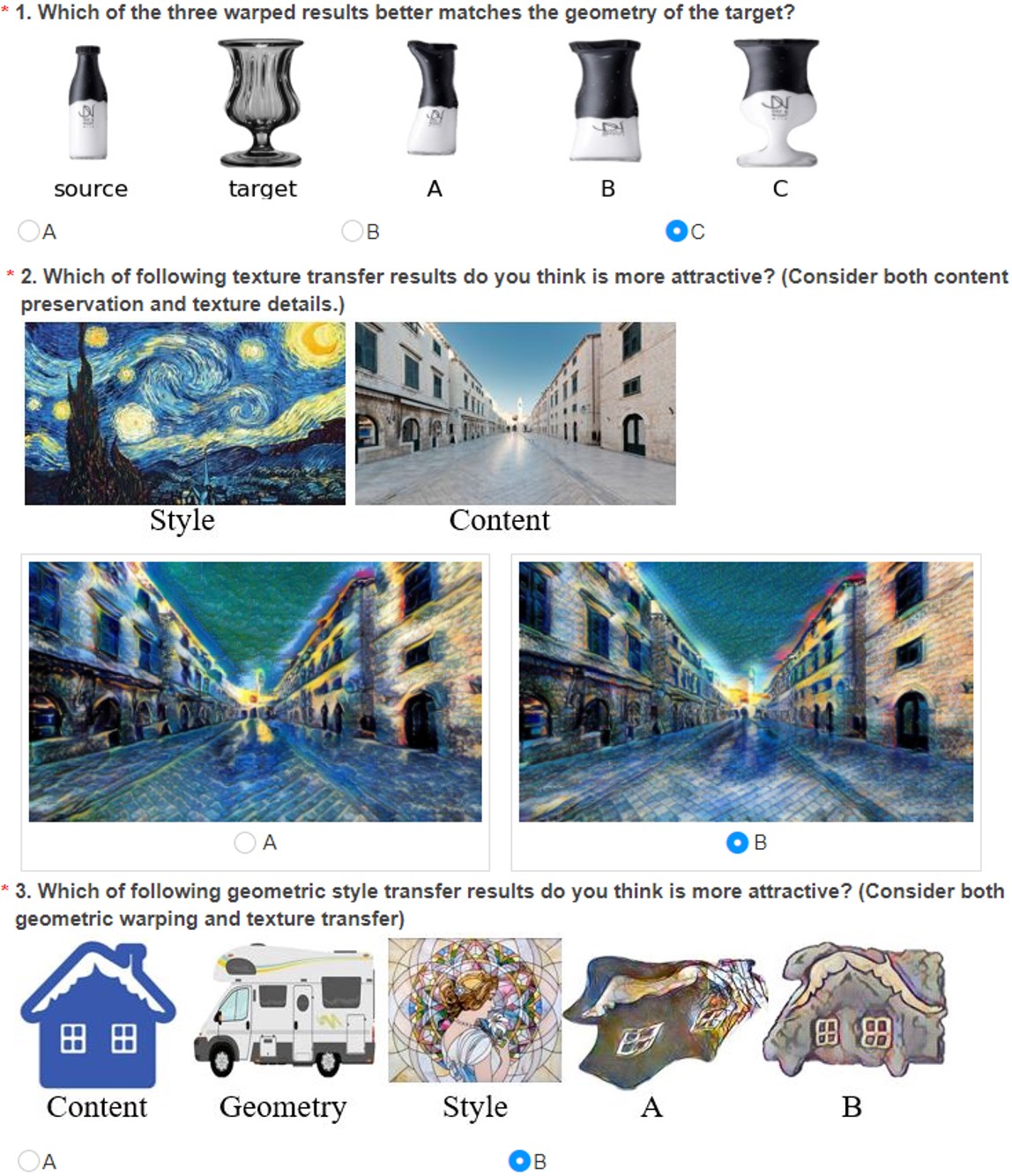}
    \vskip -0.1in
    \caption{Evaluation Interface}
    \label{fig:eval_interface}
    \vskip -0.2in
\end{figure}

As described in subsection 4.2 of the main paper, we conduct a user study to evaluate the effect of the proposed InST algorithm against the existing methods. We divide the evaluation into three groups from the perspectives of geometric warping, content maintenance, and their combination, and each group is set to ten options. The evaluation interface is partly shown in Fig.~\ref{fig:eval_interface}.

\subsection{More Results}
\label{sec:more_results}

\begin{figure*}[ht]
    \centering
    \includegraphics[width=0.90\linewidth]{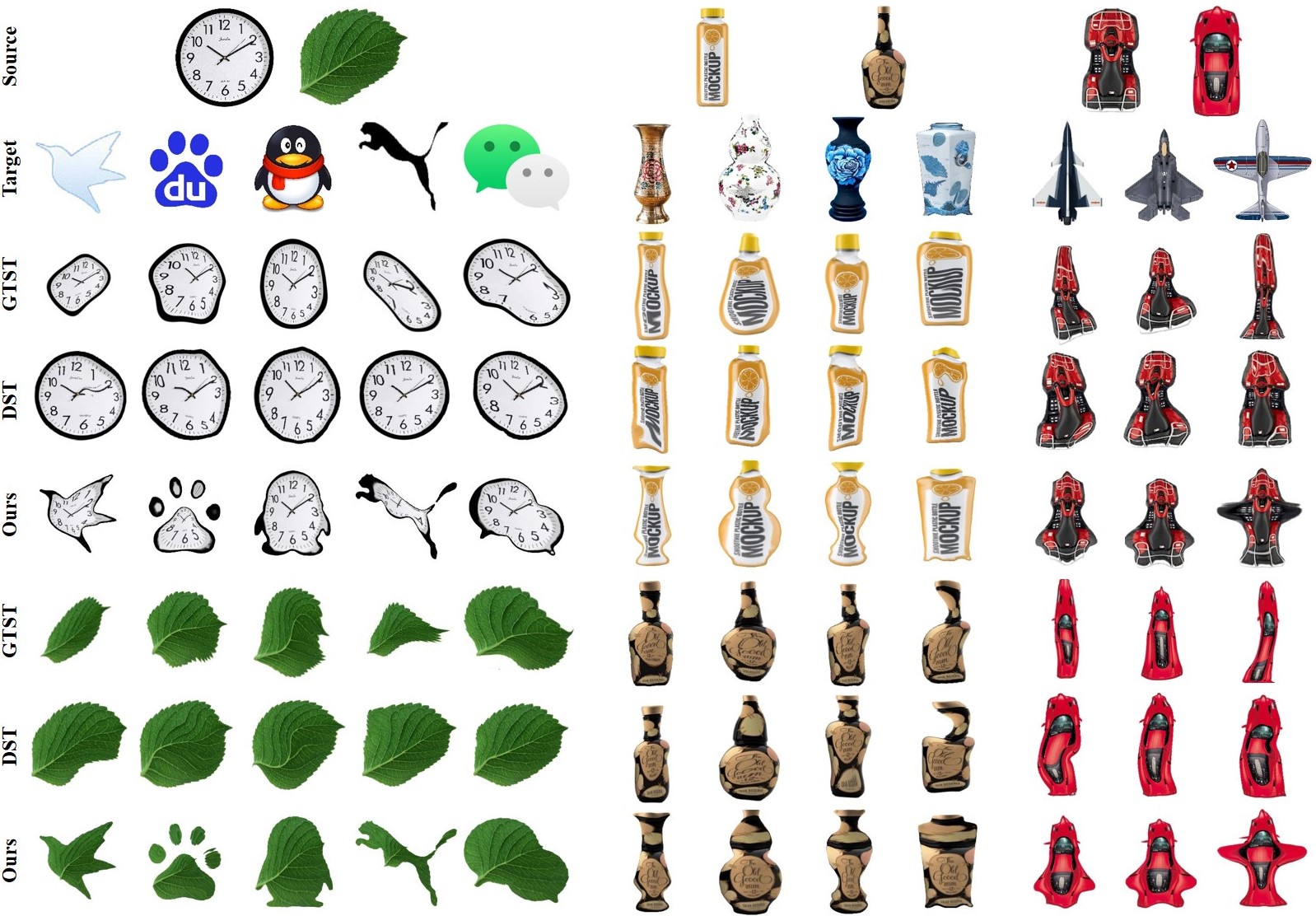}
    \vskip -0.1in
    \caption{More visual logo and product design results using the geometric transfer methods, \eg, DST \cite{Kim2020DeformableST}, GTST \cite{Liu2021warp} and our InST.}
     \vskip -0.05in
    \label{fig:warp_log}
\end{figure*}

\begin{figure*}[ht]
	\centering
	\begin{subfigure}[b]{0.9\linewidth}
		\centering
		\includegraphics[width=0.9\linewidth]{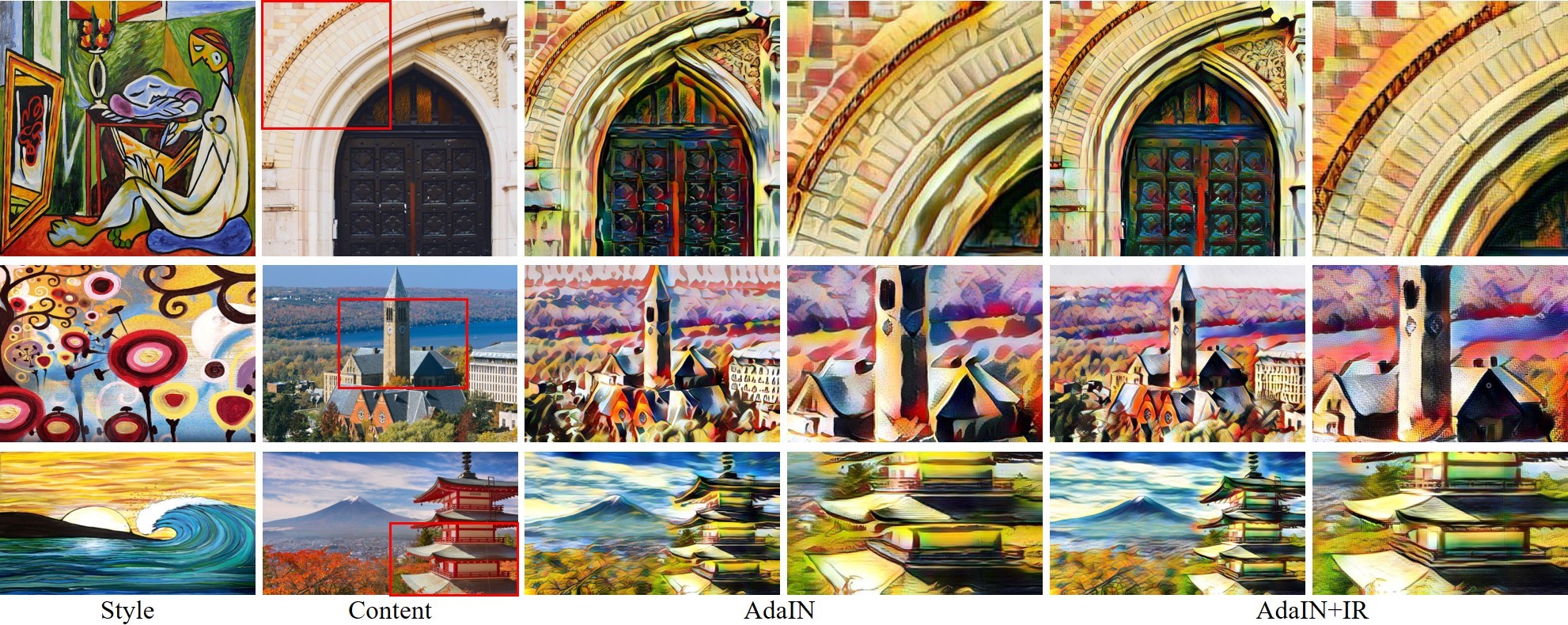}
	\end{subfigure}
	\begin{subfigure}[b]{0.9\linewidth}
		\centering
		\includegraphics[width=0.9\linewidth]{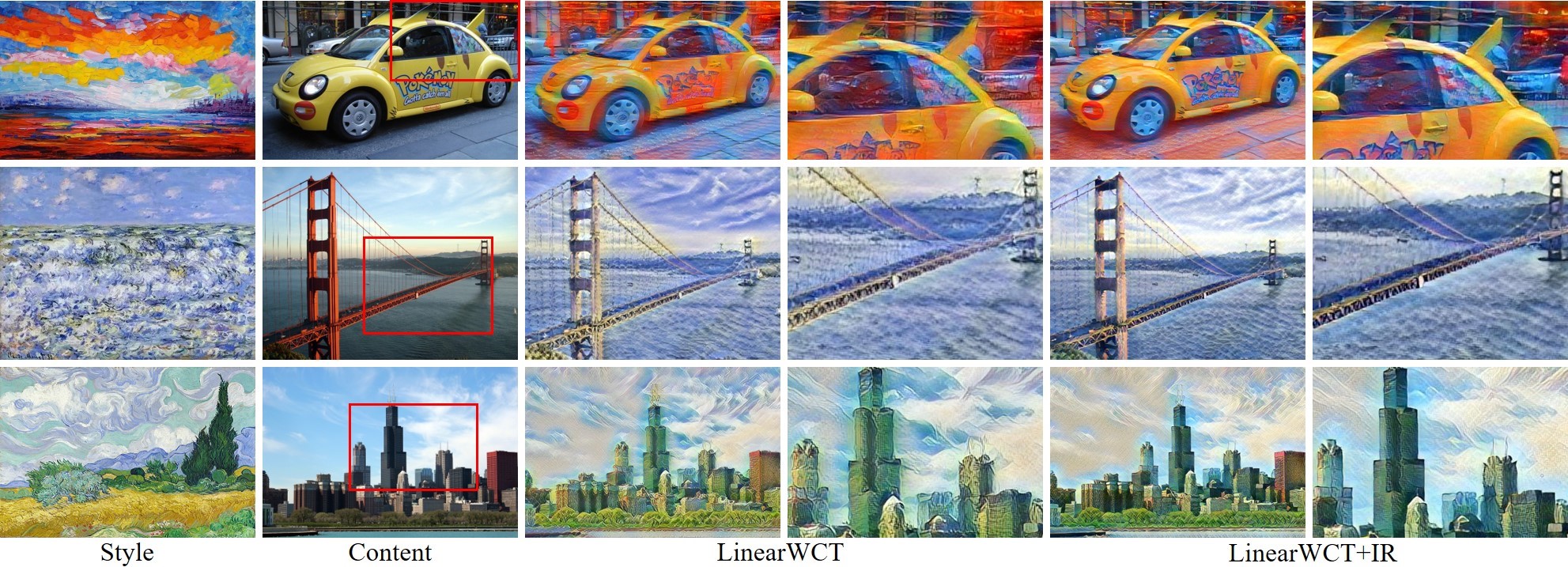}
	\end{subfigure}
	\begin{subfigure}[b]{0.9\linewidth}
		\centering
		\includegraphics[width=0.9\linewidth]{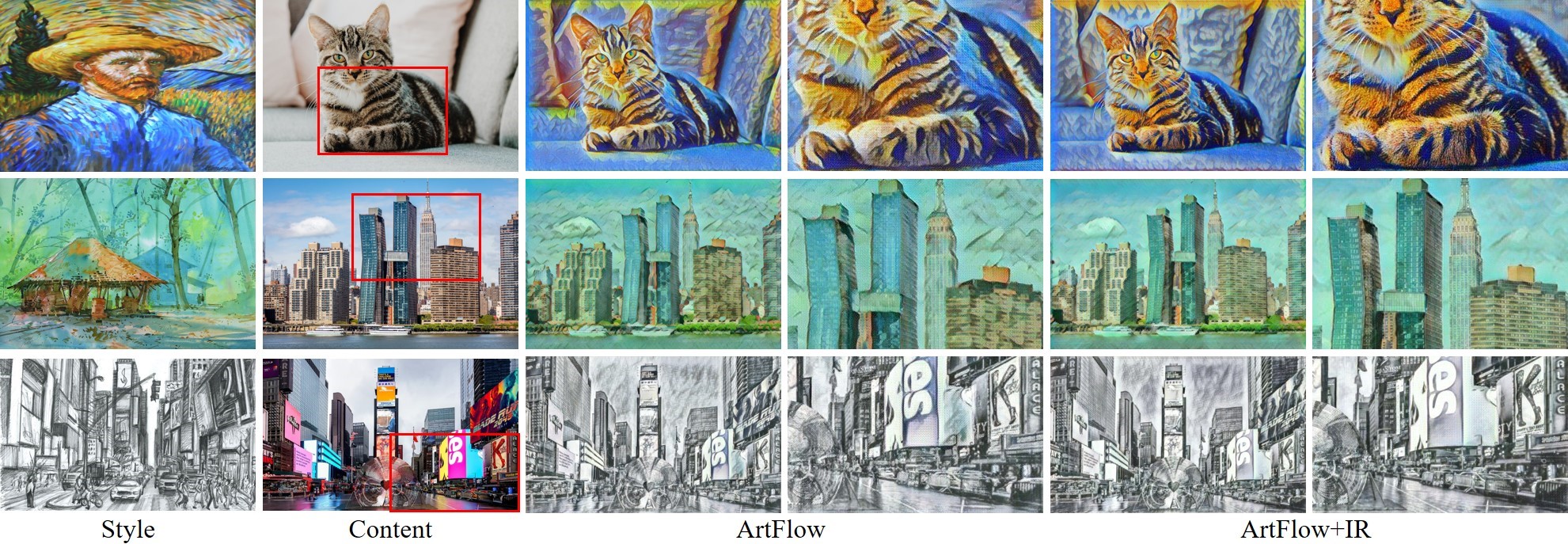}
	\end{subfigure}
	\caption{More content preservation results using the texture style transfer methods, \eg, AdaIN \cite{Huang2017adain}, LinearWCT \cite{Li2019lineartransfer} and ArtFlow \cite{An2021ArtFlow}.}
	\label{fig:NST_results}
\end{figure*}

\begin{figure*}[ht]
    \centering
    \includegraphics[width=0.90\linewidth]{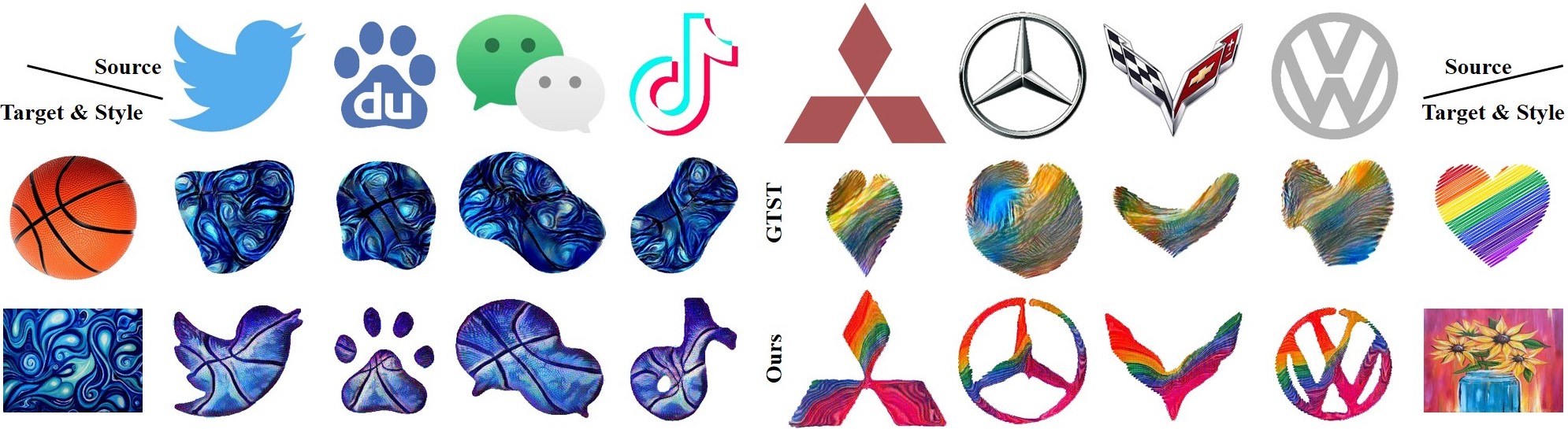}
    \vskip -0.1in
    \caption{More visual logo design results using the geometric and texture style transfer methods, \eg, GTST \cite{Liu2021warp} and our InST.}
    \vskip -0.05in
    \label{fig:combination}
\end{figure*}

\begin{figure*}[t]
    \centering
    \includegraphics[width=0.90\linewidth]{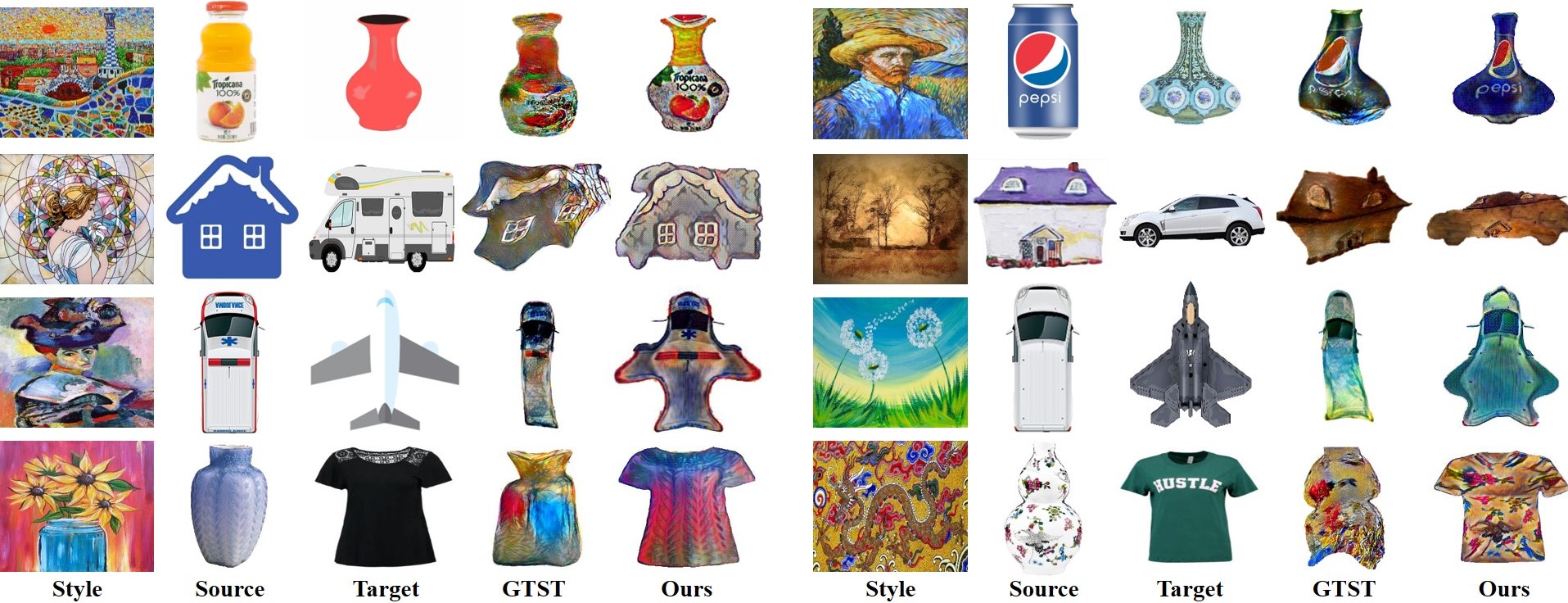}
    \vskip -0.1in
    \caption{More visual product design results using the geometric and texture style transfer methods, \eg, GTST \cite{Liu2021warp} and our InST.}
    \vskip -0.15in
    \label{fig:combination2}
\end{figure*}

Here, more qualitative results are provided to assist the readers in assessing the effectiveness of our proposed InST algorithm. We showcase the comparison results from three aspects like the main paper: (i) geometric warping (Fig.~\ref{fig:warp_log}, Fig.~\ref{fig:geometricproduct}), (ii) texture transfer (Fig.~\ref{fig:NST_results}), (iii) their combination (Figs.~\ref{fig:combination} and \ref{fig:combination2}).

\begin{figure*}[ht]
    \centering
    \includegraphics[width=0.90\linewidth]{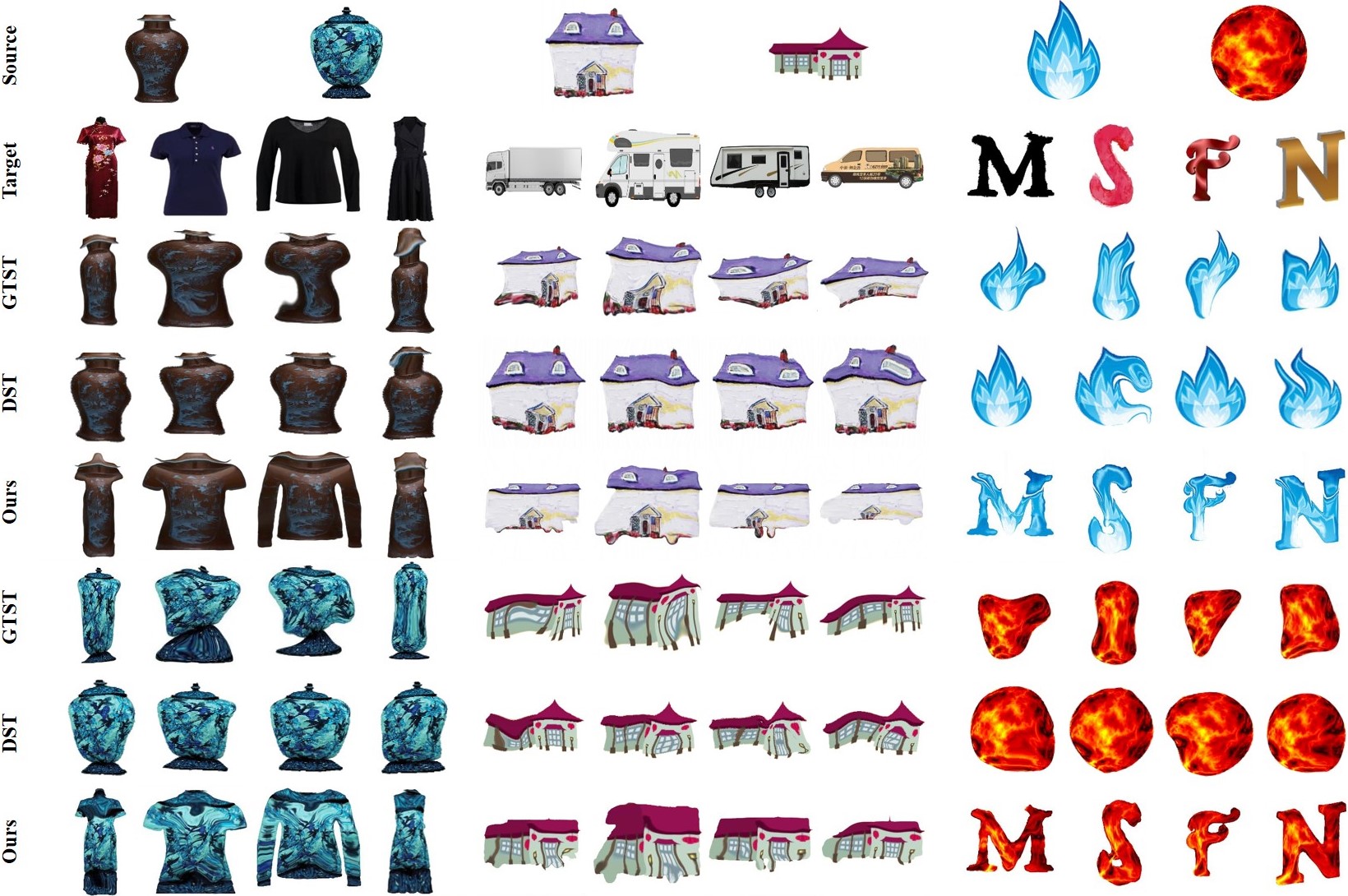}
    \vskip -0.1in
    \caption{More visual product design results using the geometric style transfer methods, \eg, DST \cite{Kim2020DeformableST}, GTST \cite{Liu2021warp} and our InST.}
     \vskip -0.05in
    \label{fig:geometricproduct}
\end{figure*}


\end{document}